\documentclass{ecai}
\usepackage{times}
\usepackage{graphicx}
\usepackage{latexsym}

\usepackage{multirow}
\usepackage{makecell}
\usepackage{tabu, array, tabularx}
\usepackage{amsmath,amssymb}
\usepackage{algpseudocode,algorithm}
\usepackage{subfig}

\ecaisubmission   % inserts page numbers. Use only for submission of paper.
\begin{document}

\title{Robust Multi-Instance Learning with Stable Instances}

\author{Weijia Zhang$^1$ \and Lin Liu$^1$ \and Jiuyong Li\institute{University of South Australia,
Australia, email: \{weijia.zhang, lin.liu, jiuyong.li\}@unisa.edu.au}}

\maketitle
\bibliographystyle{ecai}

\begin{abstract}
Multi-instance learning (MIL) deals with tasks where data is represented by a set of bags and each bag is described by a set of instances.
Unlike standard supervised learning, only the bag labels are observed whereas the label for each instance is not available to the learner.
Previous MIL studies typically follow the i.i.d. assumption, that the training and test samples are independently drawn from the same distribution. However, such assumption is often violated in real-world applications.
Efforts have been made towards addressing distribution changes by importance weighting the training data with the density ratio between the training and test samples.
Unfortunately, models often need to be trained without seeing the test distributions.
In this paper we propose possibly the first framework for addressing distribution change in MIL without requiring access to the unlabeled test data. Our framework builds upon identifying a novel connection between MIL and the potential outcome framework in causal effect estimation. 
Experimental results on synthetic distribution change datasets, real-world datasets with synthetic distribution biases and real distributional biased image classification datasets validate the effectiveness of our approach.
\end{abstract}

\section{Introduction}

Multi-instance learning (MIL) \cite{Dietterich1997} deals with tasks where the data is consisted of a set of bags and each bag contains a set of instances. Originally, MIL was proposed for drug activity prediction, where the goal is to predict whether a new molecule is qualified to make drug. Each molecule can have many different low-energy shapes; however, biochemists only know whether it is qualified for drug making at the molecule level without knowing which specific shape of the molecule is qualified. To solve this problem, \cite{Dietterich1997} proposed the multi-instance learning framework where each molecule should be modeled as a bag, and the low-energy shapes of the molecule consist its instances. 

Unlike traditional supervised learning where each sample is associated with a label, in MIL only the bag labels are available whereas instance labels are unknown. 
The relationships between the instance labels and the bag labels are defined by multi-instance assumptions. In this paper, we focus on the widely used standard multi-instance assumption \cite{Foulds2010}: a bag is labeled as positive if it contains at least one positive instance, and labeled as negative if otherwise. 
The main goal of the majority of MIL algorithms is to predict the labels of unseen test bags. Drug activity prediction asside, the problem of MIL arises naturally in many applications where label is difficult to obtain, including text categorization \cite{Andrews:2002:SVM:2968618.2968690}, web index page recommendation \cite{Zhou2005}, compute-aided diagnoses, and image classification \cite{Chen2006}.

\begin{figure*}[!t]
	\includegraphics[width=\textwidth]{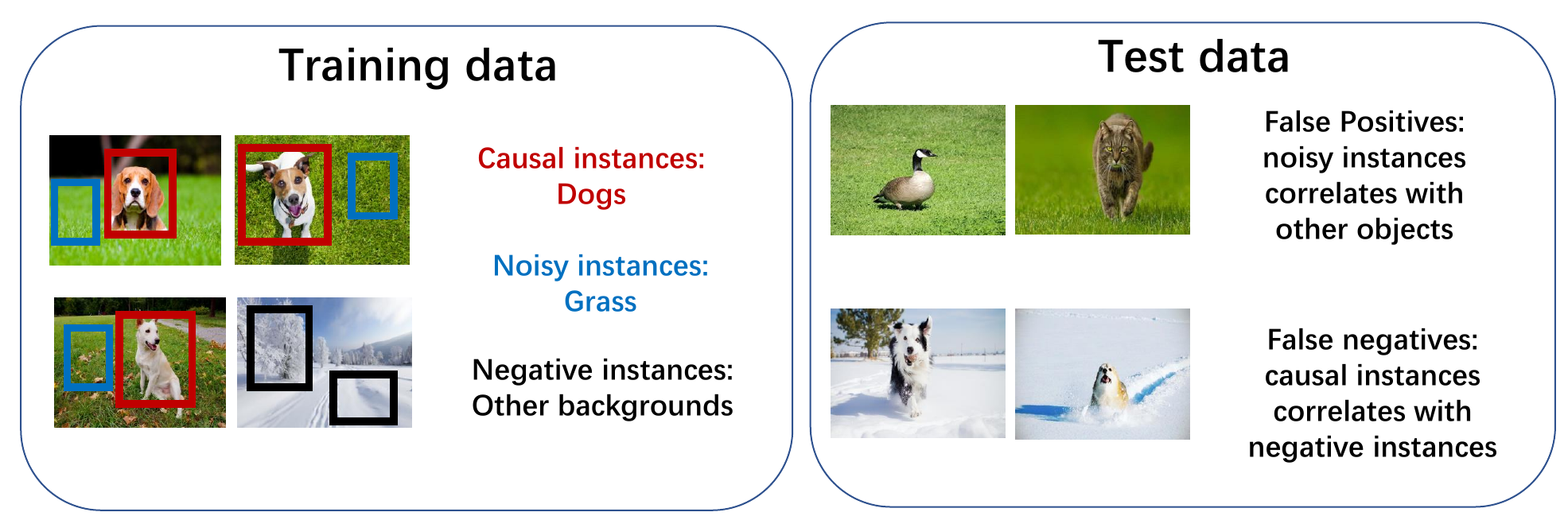}
	\caption{(Best viewed in colour) 
		Causal instances (red) are dogs. In training data, noisy instances (blue) are grass and are highly correlated with the causal instances; negative instances (black) are other random backgrounds which are negatively correlated with the causal instances. In test data, noisy instances correlate with other objects such as cats and produce false positives, causal instances correlate with negative instances and produce false negatives. 
%		When distribution change occurs, the distribution of the causal instances remains unchanged, whereas the distribution of the noisy instances changes. Existing methods do not differentiate causal instances from noisy instances, and thus their performance degenerate when the training and the test distributions differ. 
}
	\label{example}
\end{figure*}

Most multi-instance learning methods assume that the training and test data are drawn independently from an identical distribution. However, such assumption is frequently violated in real-world tasks \cite{Shimodaira2000}. Distribution change happens due to multiple reasons, i.e., when the training and test data are collected during different times or from different locations. Consider the example of an image classification task in Figure \ref{example} where the task is to train a classifier for dogs. The training images are collected during summer where the backgrounds often contain grass; however, the test samples are collected during winter where the backgrounds are mostly snowy \cite{Zhang2014}. Without accommodating the distribution change, standard supervised methods have a tendency to predict images with grass as positive and images with dogs in snow as negative because of the distribution difference between the training and test samples \cite{He2019}.

Addressing the discrepancy between training and test distributions in standard supervised learning has attracted much attention during the past decades. Many algorithms have been proposed to solve the problem \cite{BenDavid2006,Sugiyama2007}. Among these approaches the covariate shift setting, where the marginal distribution of samples changes but the conditional distribution of the class label conditioning on the samples do not change, has attracted the most attention \cite{Zadrozny2004,Sugiyama2009}. Unfortunately, most existing studies focused on single-instance setting and it has been shown that single-instance techniques for handling distribution change are not effective in multi-instance learning \cite{Zhang2014}.

Several MIL methods have been proposed to address distribution change under the covariate shift assumption by utilizing the unlabeled test data to estimate the importance weights between the test and training samples, and incorporating separate weights address the bag-level and instance-level distribution change in MIL \cite{Zhang2014,Zhang2017}. 
Unfortunately, in many real-world scenarios classifiers often need to be trained without giving access to the test samples, which renders the existing distribution change MIL methods inapplicable.

In this paper, we tackle distribution change in MIL by grouping instances into three categories: causal instances (i.e., dog in Figure \ref{example}), noisy instances (grass) and negative instances (other background instances). We assume that the probability of the bag label conditioning on the causal instances remains unchanged across the training and test data. Different from the covariate shift assumption, both the probabilities of the bag label conditioning on the noisy and negative instances, and the marginal distributions of all three types of instances can change under our assumption.

%Causal relationships describe the underlying mechanism of complex systems and thus are not affected by the distribution changes during the data collection procedure. Classifiers based on causal relationships will achieve more stable performance than correlation based methods when distribution change occurs.
%Several methods have been proposed to use causal relationships to design stable learners in single instance learning \cite{Kuang2018,Shen2018}. However, none of them are designed for multi-instance learning.

By considering adding an instance to a bag as a treatment, we propose that causal instances can be distinguished from noisy and negative instances in MIL by estimating the causal effect of an instance on the bag label under the potential outcome framework \cite{Imbens2015}. Under the standard MIL assumption \cite{Foulds2010}, adding a causal instance to a negative bag would change the label of the bag; however, the bag label would remain unchanged if the instance is noisy or negative. In other words, causal instances have greater estimated treatment effects than noisy and negative instances. Since causal instances are less affected by the distribution shift between training and test sets, using only on the causal instances will increase the robustness of a classifier.

Inspired by this motivation, we present possibly the first work, coined the StableMIL framework, to address distribution change in MIL without requiring access to the unlabeled test data. 
%StableMIL works by differentiating causal instances from noisy and negative ones 
%To the best of our knowledge, StableMIL is the first multi-instance method robust to distribution change without relying on the test distribution. 
We conduct experiments to evaluate StableMIL on synthetic datasets, real-world text and image classification tasks with synthetic distribution biases, and real-world biased image classification dataset. Results have shown that without accessing the test distribution, StableMIL significantly outperforms state-of-the-art algorithms and performs similarly to MIL distribution change methods that have access to the test data. 

The rest of this paper is organized as follows.  We review related work in Section 2, and present the proposed StableMIL framework in Section 3. Then we report the experimental results in Section 4 and we conclude the paper in Section 5.

\section{Related work}
Multi-instance learning was first proposed for drug activity prediction \cite{Dietterich1997}. Since then, many algorithms have been proposed to solve the MIL problem, which can be roughly divided into two categories: one group of methods aim to directly solve the MIL problem in either instance level \cite{Andrews:2002:SVM:2968618.2968690} or bag level \cite{Zhou2009}, and another category of methods transform MIL into single instance learning via bag embedding, among which MILES \cite{Chen2006} is a typical representative, and many methods \cite{Fu2011,Wei2017} have been proposed thereafter following this paradigm. 

Distribution change has been studied extensively in single-instance learning, among which the covariate shift setting \cite{Shimodaira2000,Huang2006} has attracted the most attention. In covariate shift, the marginal training distribution $P_{tr}(X)$ is different from the marginal test distribution $P_{te}(X)$, but the conditional distribution $P(y|X)$ remains stable across training and test samples. 
A common approach for addressing covariate shift is importance weighting, which assigns each training instance $x$ with a weight $w(x) = P_{te}(X)/P_{tr}(X)$ to diminish the discrepancy of training and test marginals \cite{Sugiyama2007}.

Distribution robust supervised learning (DRSL) has been attracting research interest during the last few years \cite{Liu2014,Hu2018}. These methods minimize the empirical risk with regard to the worst case test distribution within a specific uncertainty set considered by the algorithm. 
Unfortunately, none of them considers the setting of multi-instance learning, and directly applying single instance distribution change methods to multi-instance learning is difficult with studies shown that trivial extensions do not improve the performances \cite{Zhang2017}.
Recently, several works have proposed that building a classifier based on causal relationships lead to more stable predictions than classifiers purely based on correlations in single-instance learning \cite{Kuang2018}.

%Existing multi-instance methods for addressing distribution changes need to access of test data during the training phase \cite{Zhang2014,Zhang2017}. The main idea of these methods is utilizing probability density weighting \cite{Shimodaira2000} to balance the training samples with a density ratio so that the weighted distribution of the training bags is similar to the distribution of the test bags.

Several MIL algorithms have been proposed to address distribution change by utilizing the test distribution to calculate importance weight. MICS \cite{Zhang2014} solves the problem by modeling the bag-level and instance-level distribution changes separately and integrate the weights for training. MIKI \cite{Zhang2017} proposes that the distributions of positive instances should be modeled explicitly to address shifts of the positive concepts. 

Several algorithms tackle the multi-instance classification problem by directly identifying the positive instances in the multi-instance bags \cite{Liu2012,Ilse2018}. StableMIL can also be used to identify the positive instance; however, StableMIL is fundamentally different from previous algorithms because existing methods do not differentiate causality from association, and thus will suffer performance degeneration when the training and test distributions change. 

%For the case of unknown test data, 
%The performance of these methods relies on the diversity of their multiple training data, and they cannot address distribution shifts which do not appear in their training data. Moreover, most of these methods are highly complex, with training complexity growing exponentially with the dimension of the feature space in the worst case, which is not acceptable in high dimensional settings.

To the best of our knowledge, there is no existing work which considers the link between MIL and causal learning, and to solve the distribution change problem under multi-instance learning setting without requiring the access of unlabeled test samples during the training process.

\section{Stable Multi-instance Learning Framework}

\subsection{Notations}
Let $\mathcal{X} = \mathbb{R}^d$ denote the instance space and $\mathcal{Y} = \{0,1\}$ denote the label space. The learner is given a data set with $m$ training bags $\mathcal{B}_{tr} = \{(X_1,y_1), \cdots, (X_i,y_i),\cdots, (X_m, y_m) \}$, where $X_i = \{\pmb{x}_{i1}, \cdots, \pmb{x}_{ij}, \cdots, \pmb{x}_{x_{in_i}} \}$ is a bag of instances. For the simplicity of notation, we assume that the bags contain the same number of instances, i.e., $n_i = n$ for all $i$. Given training data $\mathcal{D}$, we denote the set containing all positive (negative) bags by $\mathcal{B}^+$ ($\mathcal{B}^-$). The number of positive (negative) bags is denoted by $m^+$ ($m^-$).
%If any positive instance exists in $X_i$, then $X_i$ is a positive bag with $y_i = +1$; otherwise $X_i$ is negative with $y_i = -1$. 

\newtheorem{assum}{Assumption}

We consider the widely accepted ``standard multi-instance learning assumption'' \cite{Foulds2010} throughout our discussion. Formally, the standard MIL assumption can be described as:
\begin{assum}
	A multi-instance bag is negative if and only if all of its instances are negative, and a bag is positive if at least one of its instances are positive: i.e.,
	\begin{equation}
	\phi(X_i) \triangleq f(h(\pmb{x}_{i1}), h(\pmb{x}_{i1}), \cdots , h(\pmb{x}_{in})),
	\label{MI Assumption}
	\end{equation}
	where $\phi$ is the bag labeling hypothesis, $f$ is the boolean OR function, and $h\in \mathcal{H}$ is a hypothesis for instances in $\mathcal{X}$.
\end{assum}
With 

%Counterfactual inference measures the causal effect of a binary treatment variable $T \in \{0,1\}$ on a outcome variable $Y$. The effect variable $Y$ has two potential outcomes: the first one is $Y(0)$ which is associated with $T=0$ when the treatment is absence, and the other one is $Y(1)$ which is associated with $T=1$ when the sample is treated. Note that in reality only one of the potential outcomes can be observed and the other cannot be observed.

\subsection{The framework}

The heart of this work is the claim that there exists a connection between multi-instance learning and randomized experiments in the causal inference literature \cite{Imbens2015}. 
Generally speaking, an experiment can be used for determining whether a causal relationship exists between a binary \emph{treatment} variable and an \emph{outcome} variable. In other words, whether the action of changing the value of the treatment would affect the value of the outcome.

With the standard multi-instance assumption, let us consider \emph{adding instance $\pmb{x}$ to a bag $X_j $} as the action of treatment and the label $Y$ of the bag as the outcome, then the causal relationship between instance $\pmb{x}$ and label $Y$ can be determined by whether the treatment changes the label from $0$ to $1$.
In other words, if instance $\pmb{x}$ has causal relationship with $Y$, adding it to a negative bag would flip the label from negative to positive. On the other hand, if instance $\pmb{x}$  is not causally related to $Y$, adding $\pmb{x}$ will not change the label. Formally we state the definition of \emph{causal instance} as:

\newtheorem{definition}[assum]{Definition}
\begin{definition}
	(\textbf{Causal instance})
	An instance $\pmb{x}_{ik}$ from a positive bag $X_i^+$ is a causal instance with regard to the bag label $Y$ if for any negative bag $X_j \in \mathcal{B}^-$, it satisfies
	\begin{equation}
	\phi^*(\pmb{x}_{ik} \cup X_j ) = 1,
	\label{causalInstance}
	\end{equation}  
	where $\phi^*$ denotes an oracle bag classifier which always return the correct label of a multi-instance bag, and $\pmb{x}_{ik} \cup X_j$ denotes a treated bag containing the instance of interest $\pmb{x}_{ik}$ along with all the instances in the pre-treatment bag $X_j$.
	\label{causalInstance}
\end{definition}

With Definition \ref{causalInstance}, we can now group the instances in a multi-instance bag into three categories: for the first category, the conditional expectation of the label has a non-zero dependence on the instances and the dependence does not change when other instances are added to the conditional set, we call these \textit{causal instances}. 
For example, dogs are causal instances of photos labeled as animals and the relationship will not change between the training and the test distributions (Figure \ref{example}).
For the second category of instances, we term them as \textit{noisy instances}. Noisy instances are correlated with either the causal instances, the bag label, or both, but do not themselves have causal relationships with the label. When conditioned on the full set of causal instances, noisy instances are independent of the bag label. For example, grasses and snows are noisy instances of an animal photo. Although they are highly correlated with the causal instances and the label, their correlations is vulnerable to changes during the collection of the training and test data.
The third category of instances are termed similar as in existing literature as \textit{negative instances}, which contains instances that are not significantly correlated to the label, i.e., random background objects without any significant correlation to the bag label.

The performances of existing MIL methods degenerate when distribution changes because they do not differentiate causal instances from noisy instances. 
When the training and test distributions differ, the correlations between noisy instances and the label will not be consistent. Therefore, models built by existing methods will be mislead by spurious correlations that only exist in the training data but are not valid in the test data.
On the other hand, a multi-instance classifier based on causal instances will achieve a more stable performance because causal relationships are not affected by the distribution changes in the training and test distributions.

\subsubsection{Learning Causal Instances from Experiment}
The key of building a distributional robust multi-instance classifier lies in differentiating causal instances from noisy ones. 
%If noisy instances are excluded, the classifier is more robust to changes between the training and test distribution due to the robustness of causal relationships.
In this section, we will discuss the identification of causal instances from an ideal experiment setting. 
%The oracle classifier correctly labels multi-instance bags
%Due to the simplicity of the discussion, for the moment we assume that we have access to an oracle classifier and discuss the case of an empirical classifier later. 
%In the experiment  between a binary treatment and outcome, in our case the treatment of interest is adding an instance to the bags, and outcome is the bag label.

By Definition \ref{causalInstance}, instances from negative bags cannot have causal relationships with the label. Therefore, we only need to consider the instances from the positive bags. Let $\bigcup \mathcal{B}^+$ denote the candidate instance pool which contains all the instances from the positive bags and $\pmb{x}_{k} \in \bigcup \mathcal{B}^+$ denote a candidate instance. For the sake of conciseness, subscript $k$ will be dropped when the context is clear. To determine whether $\pmb{x}$ is a causal instance, we need to estimate the causal effect of $\pmb{x}$ on the bag label $Y$ which can be defined as the difference between the expected label of a bag if it was treated minus the expected label of the bag if it were not treated:
%the difference between the expected label of a bag with $\pmb{x}$ added and the expected label of a bag with $\pmb{x}$ removed.
%$P(X)$ denote the original bag distribution, the estimand can be written as 
%\begin{equation}
%\tau(\pmb{x}) = \mathbb{E}_{B\sim P(X)} [F(\pmb{x}_{ik} \cup B)] - \mathbb{E}_{B \sim P(X)} [F(B)]
%\label{estimand}
%\end{equation}

\begin{equation}
\tau(\pmb{x}) = \mathbb{E}[Y(T=1)] - \mathbb{E}[Y(T=0)].
\label{TE}
\end{equation}
Here we use $Y(T=1)$ to denote the potential bag label if it were treated, i.e., the candidate instance $\pmb{x}$ is present in the bag; and we use $Y(T=0)$ to denote the potential bag label if the bag were not treated, i.e., $\pmb{x}$ is not present in the bag. 

%Two assumptions need to be satisfied for estimating Equation \ref{TE}, the stable unit treatment value assumption (SUTVA) and the ignorability assumption. The SUTVA assumption requires that treatment assignment of one sample does not affect potential outcomes of others, and treatments are stable. The ignorability assumption states that treatment assignment is independent of potential outcomes.

%The treatment assignment is randomized because for each bag there can always be a treated and an untreated version of it: one with the treatment instance and one without the treatment instance, while the other instances are exactly the same across the two bags. Furthermore, the treatment assignment is also non-interference since assigning treatment to a bag would not affect the label of any other bags.

Given a set of multi-instance bags, we can always obtain matched pairs of treated and untreated bags by adding the candidate instance $\pmb{x}$ to a bag (if $\pmb{x}$ is not in the pre-treatment bag) or removing $\pmb{x}$ from a bag (if $\pmb{x}$ is in the pre-treatment bag). 
Therefore, the causal effect can be estimated using the difference in the expectation of the realized outcomes provided by the data and the oracle classifier:
\begin{equation}
\tau (\pmb{x}) = \mathbb{E}[Y^*|T=1] - \mathbb{E}[Y^*|T=0],
\end{equation}
where $Y^*$ denotes the bag label after the treatment. Combining with the standard multi-instance assumption, we can obtain the following theorem: 
%Given the ignorability assumption, the expectation of potential bag labels can be estimated from the conditional expectation of the observed bag labels. Combined with the standard multi-instance assumption, we have the following theorem:
\begin{theorem}
	The causal effect of an instance $\pmb{x}$ on the bag label $Y$ obtained from an ideal experiment is
	\begin{align}
	\tau(\pmb{x}) = P(Y=0) \cdot \mathbb{E} [Y^*|Y=0, T=1] + const,
	\label{goal}
	\end{align}
	where const is the probability of that a bag contains one and only one positive instance with the instance being $\pmb{x}$.
	\label{unbiased}
\end{theorem}
%\begin{proof}	
\paragraph{Proof:}
%	\renewcommand{\qedsymbol}{}
	%	For the simplicity of discussion, we assume that the classes are balanced, i.e. $P(Y=1)=$ for $i \in\{0,1\}$. 
	Utilizing the tower property of conditional expectation, we have
	\begin{equation*}
	\centering
	\begin{split}
	\tau (\pmb{x})
	& = \mathbb{E}[Y^*|T=1] - \mathbb{E}[Y^*|T=0]\\
	&= \mathbb{E}[\mathbb{E}(Y^*|Y, T=1)] -\mathbb{E}[\mathbb{E}(Y^*|Y, T=0)]\\
	%\tau (\pmb{x}) &= \mathbb{E}[Y^*|T=1] - \mathbb{E}[Y^*|T=0] \\
	%	& = \frac{1}{2}(\sum\limits_{Y=0}^1 \mathbb{E}[Y^*|Y, T=1] - \sum\limits_{Y=0}^1     \mathbb{E}[Y^*|Y, T=0] ). \\
	& = \sum\limits_{i=0}^1 \mathbb{E}[Y^*|Y=i, T=1] P(Y=i) \\
	& - \sum\limits_{i=0}^1     \mathbb{E}[Y^*|Y=i, T=0] P(Y=i).
	% & = \mathbb{E}[Y^*|Y=1, T=1]P(Y=1) \\
	% & \quad + \mathbb{E} [Y^*|y=0, T=1]P(y=0) \\
	% &\quad - \mathbb{E}[Y^*|T=0,y = 1]P(y=1) \\
	% &\quad - \mathbb{E}[Y^*|T= 0, y = 0]P(y=0), \\
	\label{ATE2}
	\end{split}
	\end{equation*}
	From the Standard MI assumption, adding any instance to a positive bag or removing any instance from a negative bag will not change the bag label. Therefore we have, $\mathbb{E}[Y^*|Y=1, T=1] = 1$ and $\mathbb{E}[Y^*|Y = 0, T = 0] = 0$. Accordingly, 
	\begin{equation*}
	\centering
	\begin{split}
	\tau (\pmb{x}) & = \mathbb{E}[Y^*|Y=0, T=1]\cdotp P(Y=0) \\
	&- \mathbb{E}[Y^*|Y=1, T=0]\cdotp P(Y=1) + P(Y=1).
	\end{split}
	%	\tau (\pmb{x})  = \frac{1}{2} (\mathbb{E}[Y^*|Y=0, T=1] - \mathbb{E}[Y^*|Y=1, T=0] + 1).
	\end{equation*}
	There exists two possibilities when a positive bag is under the control treatment: the pre-treatment bag contains positive instances other than $\pmb{x}$ (may and may not contains $\pmb{x}$ itself), the bag contains $\pmb{x}$ and only $\pmb{x}$ as its positive instance. Under the standard multi-instance assumption, the expectation is $1$ and $0$ for the two scenarios, respectively.
	Let $p$ denote the probability of the second scenario, we can write $\mathbb{E}[Y^*|Y=1, T=0] = 1-p$.	
	Therefore, we have
	\begin{equation*}
	\tau (\pmb{x}) = \mathbb{E}[Y^*|Y=0, T=1]\cdotp P(Y=0) + p\cdotp P(Y=1) 
	%	\rlap{$ \Box$}
	\end{equation*} 
\newcommand*{\QEDB}{\hfill\ensuremath{\square}}%
	\QEDB
%\end{proof}

Theorem \ref{unbiased} indicates that the causal effect of an instance can be characterized by the expected treated bag label after adding the instance to a negative bag.
%The constant term in Equation \ref{unbiased} is the probability of $\pmb{x}$ being the only positive instance in the positive bags. 
It is safe to ignore the constant term during the estimation because $p$ would be small for causal instances and $p=0$ for non-causal instances. Additionally, if we assume the multi-instance bags are distributions over instances, we have $p=0$ for all instances since any $\pmb{x}$ almost surely does not appear in any particular positive bag  \cite{Doran2016}.
%Therefore, the key to identify causal instances lies in estimating the expectation term $\mathbb{E}[Y^*|Y=0, T=1]$, i.e., the expected bag label of the negative bags after the treatment. 

\subsubsection{Learning Stable Instances from Data}
Our proposed framework for stable multi-instance learning is inspired by the above discussion of learning the causal instances. However, since the oracle classifier does not exist, the instances identified using a surrogate empirical classifier may not be causal. We hence refer to them as stable instances to avoid misconceptions.

To identify the stable instances, first we train a multi-instance classification algorithm $\mathcal{A}$ with the training data and use $A$ to denote the classifier that $\mathcal{A}$ returns. For each candidate instance $\pmb{x}$, we then construct a set of bags which contains $m^-$ number of ``treated'' bags from the original negative bags. The treated bags are constructed by adding the candidate instance $\pmb{x}$ into the negative bags $X_i^-$ as $ X^{\pmb{x}}_i = \pmb{x} \bigcup X^{-}_{i}$, for $i = 1,\cdots,m^-$. For each treated bag, we use the previously trained classifier $A$ to predict its label. Finally we use the average of predicted labels $A(X_i^{\pmb{x}})$ on the treated bags to estimate the expectation term in Equation \ref{goal} as 
\begin{equation}
\hat{\tau}(\pmb{x}) = \frac{1}{m^-}\sum\limits^{m^-}_{i=1} A(X^{\pmb{x}}_i).
\label{empirical}
\end{equation}
After Equation \ref{empirical} has been estimated for all the candidate instances, we choose the instances with a score $s$ higher than $\tau$ as stable instances. Note that here we only select the stable instances to be included in $\mathcal{C}$ and we exclude all negative instances. Although it has been shown that in i.i.d. MIL negative instances can be helpful to the learner \cite{Fu2011}, in distribution change negative instances can still be misleading since the causal instances may correlate with negative instances in the test data.
We summarize the procedure of StableMIL for learning the stable instances in Algorithm \ref{pseudocode}.

%It is worth noting that when discussing the causal effects in Definition \ref{causalInstance} and Theorem \ref{unbiased}, we have assumed the existence of an oracle classifier which provides the ground truth label for the treated negative bags. However, such oracle would not be accessible in the real-world. We have learned a multi-instance classifier using the training samples in StableMIL. Therefore the instances in $\mathcal{C}$ are  \emph{not guaranteed} to be causally related to the label, since the bag labels are obtained by prediction instead of observation. 
%As a result, the causal relationships between instances in $\mathcal{C}$ with the label cannot be guaranteed. 

%Even when viewed from a non-causal perspective, the proposed StableMIL framework is more robust to distribution change than existing MIL methods. In the example of Figure \ref{example}, based on the training data classifier $A$ will predict images with dog in grass and some images with only grass as positive, and will predict some images with dogs in snow as negative. 
%In Algorithm \ref{pseudocode}, grass instances would have a lower probability of flipping the label when added to the negative bags than the dog instances, and thus grass instances will be excluded from $\mathcal{C}$. The test bags can then be classified according to their similarities with the instances in $\mathcal{C}$. Because grass instances are not included in $\mathcal{C}$, images containing dogs in snow are more likely to be correctly classified by StableMIL.

Even when considered from a non-causal perspective, StableMIL is more robust to distribution change than standard MIL methods. Let us assume that the training bags in Figure \ref{example} consist 40\% of images with dog in grass background, 10\% of images with dog in snow background, 10\% of grass images, 10\% of snow images, and the rest are other generic negative images.
In the procedure of Algorithm \ref{pseudocode}, grass and snow instances will have a lower chance of flipping the label when added to  negative bags than the dog instances: dog instances exist in 100\% of positive bags; however, grass instances only exist in 80\% of the positive bags but also exist in 20\% of the negative bags, and snow instances only exist in 20\% of the positive images but also exist in 20\% of the negative ones. Therefore, stable instances can be differentiated from noisy instances.

\begin{algorithm}[!t]
	\caption{Stable Multi Instance Learning}\label{euclid}
	\begin{algorithmic}[1]
		\Require{Training bags $\mathcal{B}_{tr} = \{(X_i^{tr},y_i)\}_{i=1}^m$, multi-instance algorithm $\mathcal{A}$, threshold $\tau$} 
		\Ensure{Stable Instance Pool $\mathcal{C}$}
		%		\Procedure{Euclid}{$a,b$}\Comment{The g.c.d. of a and b}
		\State Train $\mathcal{A}$ using data in  $\mathcal{B}_{tr}$
%		\State $X^+ \gets \emptyset$
		\For{every $\pmb{x} \in \cup\mathcal{B}^+$}
%		\State $X^+ \gets X^+ \cup \pmb{x} $
%		\EndFor
%		\For{every $\pmb{x} \in X^+$}
		\State $s \gets 0$
		\For{every $X_i$ in $X^-$}
		\State $s = s + A(\pmb{x} \cup X_i^-)$
		\EndFor
		\State $s \gets s /  m^-$ 
		\If{$s \geq \tau$}
		\State $\mathcal{C} = \pmb{x} \cup \mathcal{C}$
		\EndIf
		\EndFor
		\For{every $X \in \mathcal{B}_{tr}$}
		\For{every $\pmb{x} \in \mathcal{C}$}
		\State $d(X_i, \pmb{x})= \max_{\pmb{x}_{ij}} \exp ( -\lambda \Vert \pmb{x}_{ij} -\pmb{x} \Vert^2)$
		\EndFor
		\State $\pmb{z}_i = [d(X_i, \pmb{x}_{1}), \cdots, d(X_i, \pmb{x}_{q})]$, $\pmb{x}_1, \cdots,\pmb{x}_q \in \mathcal{C}$
		\EndFor
		\State Train classifier $\mathcal{L}$ with the embedded feature vectors $\pmb{z}_i$
		\State \textbf{return} $\mathcal{L}$
	\end{algorithmic}
	\label{pseudocode}
\end{algorithm}
\subsubsection{Bag Embedding and Embedded Classification}
After learning the stable instance set $C$, we used bag embedding \cite{Chen2006,Fu2011,Zhang2017} to map the bags into single instance representation using the stable instances, and then construct our classifier. 
%Bag embedding based algorithms have been very successful for solving multi-instance problems. By using b

The multi-instance bag embedding is performed based on the similarity between the bags and each of the instances in $\mathcal{C}$. The similarity between a bag and an instance can be measured using the following function:
\begin{equation}
d(X_i, \pmb{x}) = \max\limits_{\pmb{x}_{ij} \in X_i} \exp ( -\lambda \Vert \pmb{x}_{ij} -\pmb{x} \Vert^2)
\end{equation}
where $\lambda$ is the scaling parameter and can be chosen automatically by local scaling \cite{Zelnik-Manor2004}.
The intuition is that positive bags should have high similarities with at least one instance in $\mathcal{C}$, and negative bags should have low similarities with all instances in $\mathcal{C}$.
%since all instances in the negative bags are representing different concepts from the stable instances.

%Concretely, each bag is mapped into a $n$ dimensional vector $\pmb{v}_i$, and the $j$-th attribute $v_{ij}$ is 
%\begin{equation}
%v_{ij} = \max\limits_{\pmb{x}_{ik} \in X_i } \text{exp}(-\gamma || \pmb{x}_{ik} - C_j ||^2 ),
%\end{equation}
%where $C_j$ is the $j-th$ instance of $\mathcal{C}$. The embedded representations keep both bag-level information and causal instance information.

As a result, the embedded feature vector for bag $X_i$ is a $q$-dimensional vector which is consisted of the concatenated bag-to-instance similarities:
\begin{equation}
\pmb{z}_i = [d(X_i, \pmb{x}_{1}), \cdots, d(X_i, \pmb{x}_{j}), \cdots, d(X_i, \pmb{x}_{q})],
\label{mapping}
\end{equation}
where $\pmb{x}_{j} \in \mathcal{C}$ and $q$ is the cardinality of set $\mathcal{C}$. 
After the bag embeddings are calculated, any standard machine learning algorithm can be used for training the model. In our experiments, we use SVM with RBF kernel for fairness of comparison since it is also used as the classifier for many multi-instance algorithms based on the idea of bag embedding.

\setlength{\tabcolsep}{5pt}

\begin{table*}[!t]
	\centering
	\caption{Testing accuracy (\%, mean $\pm$ std.) on synthetic datasets. The highest average accuracy is marked in bold. 
		%		The last row shows the win/tie/loss counts of StableMIL versus other methods (paired \textit{t}-tests at 95\% significance level).
	}
	\begin{tabular}[width = \textwidth]{ l| l l l l | l l | l }
		\hline
		\multirow{2}{*}{} & \multicolumn{4}{c|}{State-of-the-art multi-instance learning methods} & \multicolumn{2}{c|}{\makecell{Distribution change\\ multi-instance methods}} & \makecell{Proposed \\ method}\\
		\cline{2-8} & miSVM & MILES & miGraph & miFV  & MICS  & MIKI & StableMIL \\
		\hline
		
		Setting 1 & 79.8 \small{$\pm$ 2.0} & 83.6 \small{$\pm$ 1.6} & 90.7 \small{$\pm$ 3.9} & 76.4 \small{$\pm$ 3.8}   & 91.4 \small{$\pm$ 4.9} & 87.3 \small{$\pm$ 2.5} & \textbf{93.9 \small{$\pm$ 2.3}}\\
		\hline
		Setting 2 & 72.7 \small{$\pm$ 2.5} &  71.0 \small{$\pm$ 2.8} & 70.7 \small{$\pm$ 3.0} & 68.3 \small{$\pm$ 3.6}   & 70.5 \small{$\pm$ 5.9} & 72.9 \small{$\pm$ 2.3} &  \textbf{78.9 \small{$\pm$ 3.5}} \\
		\hline
	\end{tabular}
	\label{simulation}
\end{table*}

\subsubsection{Time Complexity Analysis}
StableMIL is scalable because the identification of stable instances only requires
predicting the label of the constructed treated bags, while the training of bag-level classifier only uses the original multi-instance bags. 
We include a detailed complexity analysis as follows. 
%Suppose that we have $m$ training bags where the dimensionality of an instance is d. 
Without loss of generality and for the simplicity of analysis, we assume that each bag has the same number of $n$ instances and the number of positive/negative bags are balanced. In the training phase, StableMIL first trains a multi-instance bag classifier $A$ using the original training bags. Suppose that we use miFV as the bag classifier \cite{Wei2017}, the complexity of $O(nmd)$ plus the cost of training a linear SVM. Then we use the trained miFV model to predict the label of the constructed bags, for which the complexity is $O(n^2md)$. Note that here the quadratic term only depends on the number of instances $n$ which is usually significantly smaller than the number of bags $m$, and the predictions can be easily parallelized. Afterwards, $q$ stable instances are selected and the bag-mapping is performed with regard to the stable instances with complexity of $O(qnmd)$. Now the remaining computation for training StableMIL is training a SVM using $n$ sample of $q$ dimensional feature vectors. To sum up, the time complexity of StableMIL is $O(n^2md)$ plus training cost of two SVMs.

%Since the feature mapping defined in Equation \ref{mapping} is non-linear due to the use of exponential function, a linear classifier will suffice to separate the features.
%In this paper, we use Support Vector Machine to demonstrate the effectiveness of the selected causal instances.

\section{Empirical validation}
In this section, we empirically validate the performance of StableMIL. Firstly, we compare StableMIL with benchmark multi-instance algorithms including miSVM \cite{Andrews:2002:SVM:2968618.2968690}, MILES \cite{Chen2006}, miGraph \cite{Zhou2009}, and miFV \cite{Wei2017}.
Secondly, we compare with distribution change MIL methods that utilize the test distribution, including MICS \cite{Zhang2014} and MIKI \cite{Zhang2017}. 
Hyper parameters for the compared methods are selected either as the default recommended by the original authors (for MIKI) or using cross-validated grid search on the training samples (for other algorithms).
% The code for performing the experiments are included in the supplemen

We use miFV as the base classifier of StableMIL and SVM with Gaussian kernel for the classification after instance embedding. Although using instance level classifiers is more theoretical consistent with the standard MI assumption. However, since it has been shown that instance-level MIL methods generally perform significantly worse than bag-level methods \cite{Amores2013}, we choose to use a state-of-the-art bag-level classifier miFV because we want the base classifier to be as accurate as possible.

We use SVM with Gaussian kernel as the classifier of StableMIL after the multi-instance bags has been transformed to a single instance after stable instance embedding. A reason for choosing Gaussian SVM is because it is also used in the compared methods (except miFV which uses linear SVM). Thus, it can be seen that the improvement is due to identification of the causal instances by StableMIL. We tested using linear kernel instead of Gaussian, the trend remains similar, but the accuracies are slightly lower than Gaussian for all compared algorithms.

%If we know the expected number of causal instances in each bag, $\tau$ can be chosen to ensure the number of instances in $\mathcal{C}$ contains around the same number of causal instances.
For choosing the stable instances threshold $\tau$ of StableMIL, we first split the negative bags in the training data into two equally-sized parts. Then we construct a set of treated bags by adding the instances from the first part to the bags from the second part of the negative training bags. After estimating the scores $s_{neg}$ for the constructed bags, we use the third quartile value of $s_{neg}$ for parameter $\tau$. Other parameters for StableMIL are selected as same as those of miFV.

%The parameter $\tau$ of StableMIL is selected to ensure that the cardinality of stable instance pool $\mathcal{C}$ is larger than the number of positive bags. 

%We use miFV as the base multi-instance learner and SVM with linear kernel as the single instance learner to instantialize StableMIL. The parameter of these two learners are fixed across all datasets for StableMIL. Since the threshold for causal instance pool $\tau$ is adaptively selected using the strategy mentioned in the last section, no parameter tuning is performed for StableMIL. For compared methods, the parameter is selected as the better one of the following two: the default recommended in the original paper and the parameter selected via 5-fold cross-validation on the training data.

\subsection{Synthetic Data}
\label{synthetic}
%We first evaluate StableMIL with synthetic data and bias. 
%The positive, noisy and negative instances are generated from three different multi-variate Gaussian distributions. 
%For the training data most positive bags contain positive instances together with noisy instance while the negative bags contains mostly negative instances. 
%For example, if we consider the positive instances as the dog concept, noisy instances as the grass concept, and negative instances as the snow concept, then the positive bags in the training data are mostly dog on grass pictures, while the negative bags are mostly snowy pictures. In the test data, dog concept would be associated with snow in the positive bags, while negative bags contain many grass images.
%%The positive concepts are generated from two normal distributions, and the negative concepts are generated from another two different distributions. 
%
%Specifically, each bag contains a random number of $10$ to $20$ instances. On average each positive bag contains 2 positive instances. A total number of 200 positive bags and 200 negative bags are generated. We vary the dimensions of the Gaussian distributions and generate two settings with 5 and 100 dimensions, respectively.
We first evaluate StableMIL with synthetic data. 
The instances in the multi-instance bags are generated by sampling from four distinct multivariate Gaussian distributions: the first one for the causal instances, the second one for the noisy instance, the third one and the fourth one for the negative instances. 

For example, let us suppose the task is to classify dog and the four distributions describe ``dog'', ``grass'', ``snow'' and ``cat'', respectively. 
In the training data most of the positive bags describe ``dog on grass'' with only a small number of bags describe ``dog on snow''; for the negative bags, most of them describe ``snow'' and the rest are ``grass'' or ``cat''. In the test data, the bias is reversed: most positive bags are ``dog on snow'' while the majority of negative bags are ``grass'' or ``cat on grass''.

%Each bag contains around $10$ to $20$ instances. Each positive bag contains an average of 2 positive instances which comes from one of the two positive distributions, and each negative bag contains instances from one of the two negative distributions. A total number of 200 positive bags and 200 negative bags are generated with positive instances from each concept distributed evenly. We vary the dimensions of the distributions to generate low and high dimensional settings.

More specifically, the distribution changes between the training and test datasets are generated using a biased sampling procedure similar to those used in relevant literature \cite{Zadrozny2004,Zhang2014,Zhang2017}. We define a selection variable $s_i$, where $s_i = 1$ indicates the $i$-th bag is selected into the training set and $s_i=0$ indicates the bag falls into the test set. Denote the positive and negative instances as $\pmb{x}^+$ and $\pmb{x}^-$, then for positive bags the sampling rules are
\begin{align*}
&Pr(s_i = 1 | \pmb{x}^+ \in \mathcal{P}, \pmb{x}^- \in \mathcal{N}_1) = a.\\ 
&Pr(s_i = 1 | \pmb{x}^+ \in \mathcal{P}, \pmb{x}^- \in \mathcal{N}_2) = 1-a.
\end{align*}
For negative bags the rules are
\begin{align*}
&Pr(s_i = 1 | \pmb{x}^- \in \mathcal{N}_1) = 1-a \\
&Pr(s_i = 1 | \pmb{x}^- \in \mathcal{N}_j) = a. \text{ for } j=2,3
%&Pr(s_i = 1 | \pmb{x}^- \in \mathcal{N}_3) = a.
\end{align*}
Here $\mathcal{P}_1$, $\mathcal{P}_2$, $\mathcal{N}_1$ and $\mathcal{N}_2$ are multivariate Gaussian distributions where the positive and negative instances are sampled from. 
%Specifically, we set $P_i=\mathcal{N}(\mu_i,\sigma)$ with $\mu_i = (i,i)$ and $\sigma^2 = (0.5,0,5)$. 
The values of sampling ratio $a$ are uniformly selected from $a\in [0.65,0.95]$ at each simulation. Two different settings of simulated datasets are generated and they differ in the dimensionality of the instances. In Setting 1 each instance is a 3-dimensional feature vector while the instances in Setting 2 are 100-dimensional feature vectors.
%All datasets and scripts used in the experiments can be accessed online.\footnote{https://www.dropbox.com/s/lxi083v9l8kmuxr/Supplementary.zip?dl=0} 

We repeated experiments for 30 times by using the sampling procedure to generate training and test sets and report the simulation results in Table \ref{simulation}. Firtly, the proposed StableMIL method achieves significantly better performance than the compared i.i.d. MIL methods. 
Secondly, when compared with distribution change based MIL algorithms which has access to the unlabeled test samples, StableMIL also achieves better performance without seeing the samples.

\subsection{MNIST Dataset}

\begin{table*}[!t]
	\centering
	\caption{Testing accuracy (\%, mean $\pm$ std.) on MNIST. The highest average accuracy is marked in bold. $\bullet$/$\circ$ indicates that StableMIL is significantly better/worse than the compared methods (paired \textit{t}-tests at 95\% significance level). The last row summarizes the Win/Tie/Lose counts of StableMIL versus other methods.}
	\begin{tabular}{ c | l l  l l | l l | l }
		\hline
		\multirow{2}{*}{} & \multicolumn{4}{c|}{State-of-the-art multi-instance learning methods} & \multicolumn{2}{c|}{\makecell{Distribution change\\ multi-instance methods}} & \makecell{Proposed \\ method}\\
		\cline{2-8}
		& miSVM & MILES & miGraph & miFV  & MICS  & MIKI & StableMIL \\
		\hline
		
		6 and 9 & 59.3 \small{$\pm$ 10.3}$\bullet$ & 77.7 \small{$\pm$ 7.6}$\bullet$ & 78.5 \small{$\pm$ 10.8}$\bullet$ & 70.8  \small{$\pm$ 12.7}$\bullet$  & 80.0 \small{$\pm$ 10.2}$\bullet$  & \textbf{92.8 \small{$\pm$ 7.9}} & 91.9 \small{$\pm$ 7.1}\\
		
		0 and 6 & 54.9 \small{$\pm$ 7.1}$\bullet$ & 79.2 \small{$\pm$ 10.6}$\bullet$ & 75.5 \small{$\pm$ 11.5}$\bullet$ & 75.0 \small{$\pm$ 12.8}$\bullet$  & 75.6 \small{$\pm$ 11.3}$\bullet$  & \textbf{94.6 \small{$\pm$ 4.3}}$\circ$ & 93.3 \small{$\pm$ 4.6}\\
		
		0 and 8 & 53.6 $\pm$ \small{9.8}$\bullet$ &  72.8 $\pm$ \small{9.5}$\bullet$ & 74.3 $\pm$ \small{10.9}$\bullet$ & 70.0  \small{$\pm$ 15.0}$\bullet$  & 75.4 \small{$\pm$ 11.2}$\bullet$ & \textbf{94.7 \small{$\pm$ 4.3}}$\circ$ & 91.8 \small{$\pm$ 8.8}\\
		
		0 and 9 &  56.9 \small{$\pm$ 10.4}$\bullet$ & 71.9 \small{$\pm$ 10.0}$\bullet$ & 74.6 \small{$\pm$ 11.0}$\bullet$  & 70.5 \small{$\pm$ 10.8}$\bullet$ &   74.0 \small{$\pm$ 10.3}$\bullet$  & 95.4 \small{$\pm$ 2.5}  & \textbf{95.8  \small{$\pm$ 3.3}}\\
		
		4 and 7 & 61.4 \small{$\pm$ 13.0}$\bullet$ & 68.9 \small{$\pm$ 10.5}$\bullet$ & 74.5 \small{$\pm$ 11.3}$\bullet$ & 71.7 \small{$\pm$ 13.9}$\bullet$ &   74.6 \small{$\pm$ 10.7}$\bullet$  & \textbf{94.8 \small{$\pm$ 7.7}} & 94.6 \small{$\pm$ 6.2}\\
		
		1 and 7 & 58.4 \small{$\pm$ 11.2}$\bullet$ & 90.5 \small{$\pm$ 7.1}$\bullet$ & 77.5 \small{$\pm$ 11.3}$\bullet$  & 76.0 \small{$\pm$ 10.8}$\bullet$ &  70.0 \small{$\pm$ 11.6}$\bullet$  & 92.4 \small{$\pm$ 5.7}$\bullet$ & \textbf{95.4 \small{$\pm$ 5.0}}\\
		
		2 and 7 & 57.9 \small{$\pm$ 11.3}$\bullet$ & 59.3 \small{$\pm$ 5.9}$\bullet$ & 75.2 \small{$\pm$ 12.0}$\bullet$  & 70.4 \small{$\pm$ 13.8}$\bullet$ &   78.6 \small{$\pm$ 10.2}$\bullet$ & \textbf{92.1 \small{$\pm$ 10.4}} & 90.1 \small{$\pm$ 5.6}\\
		
		3 and 6 & 57.6 \small{$\pm$ 8.9}$\bullet$ & 61.6 \small{$\pm$ 8.1}$\bullet$ &  77.6 \small{$\pm$ 12.2}$\bullet$ & 68.8 \small{$\pm$ 10.9}$\bullet$ &  76.1\small{$\pm$ 10.2}$\bullet$& \textbf{95.0 \small{$\pm$ 3.0}}$\circ$ & 92.6 \small{$\pm$ 7.9}\\
		
		6 and 8 & 57.9 \small{$\pm$ 11.3}$\bullet$ & 76.0 \small{$\pm$ 7.7}$\bullet$ & 76.3 \small{$\pm$ 11.3}$\bullet$ & 71.0 \small{$\pm$ 14.3}$\bullet$ &  74.1 \small{$\pm$ 10.5}$\bullet$  & \textbf{92.7 \small{$\pm$ 5.9}} & 92.1 \small{$\pm$ 5.6} \\
		
		2 and 4 & 58.5 \small{$\pm$ 8.9}$\bullet$ & 60.0 \small{$\pm$ 6.6}$\bullet$ & 72.4 \small{$\pm$ 10.3}$\bullet$  & 71.7 \small{$\pm$ 12.2}$\bullet$  & 74.1 \small{$\pm$ 11.8}$\bullet$ &  \textbf{94.7 \small{$\pm$ 3.4}$\circ$} & 92.0 \small{$\pm$ 6.5} \\	
		\hline
		W/T/L & 10/0/0 & 10/0/0 & 10/0/0 & 10/0/0 & 10/0/0 & 1/5/4 \\
		\hline
	\end{tabular}
	\label{image}
\end{table*}
Next we evaluate StableMIL for distribution change in image classification using the multi-instance MNIST dataset. 
%Each sample in MNIST is an image of a hand-written digit of zero to nine.  
We generate 200 positive and 200 negative multi-instance bags where each bag contains an average of 20 instances (digits). The feature for each instance is extracted using LeNet-5 \cite{Lecun1998}.
A bag is labeled as positive if it contains a specific digit (i.e., ``1''), we use a visually similar digit as an analogy of the ``cat'' negative concept (i.e., ``1'' and ``7''), and separate the rest of the digits into two groups as the other negative concepts.
The biased sampling procedure is performed similarly to the last section.
The reported results are averaged over repeating the biased sampling procedure for 30 times.

%We randomly separate the non-positive digits into two disjoint subgroups and sample the negative instances from each of them. For example, when compared to the test data, in the training data there are more positive bags with digit ``1'' as positive instances and ``2'', ``5'', ``6'', ``9'' as negative instances but less positive bags with ``7'' and ``3'', ``4'', ``8'', ``0'', and there are more negative bags with ``3'', ``4'', ``8'', ``0'' as instances but less negative bags with ``2'', ``5'', ``6'', ``9''. 

%Half of the positive bags contains positive instances from one sub-concept (e.g., `1'), and the other half from another sub-concept (e.g., `7').  
%Each positive bag contains an average of 1.5 positive instances, and the negative instances are sampled from two groups of negative classes. 

%In the negative bags, there are more positive bags with `7' as positive and `3', `5', `9', `0' as negative instance

We report the results Table \ref{image}.
When the distribution bias between training and test samples exists, the performance of StableMIL is significantly better on all datasets when compared with state-of-the-art MIL algorithms. Moreover, it can be seen that the standard deviation of the accuracies in StableMIL is much lower than the compared methods. This indicates that StableMIL is less susceptible to the variations between the training and test distributions.

If we consider the fact that MILES is a bag embedding based MIL algorithm using the union of the instances in positive bags and StableMIL performs bag embedding using only the stable instances, the superior performance of StableMIL over MILES indicates that the stable instances selected by StableMIL are indeed useful for stable prediction across unknown test distributions.

When compared to distribution change methods, the performance of StableMIL is similar to MIKI (Wilcoxon rank-sum test at $p=0.05$ indicates the performances of StableMIL and MIKI are not statistically significant); however, MIKI needs to access the test distribution whereas StableMIL achieves competitive accuracy using only the training data. 

\begin{table*}[t]
	\centering
	\caption{Testing accuracy (\%, mean $\pm$ std.) on 20 Newsgroup. The highest average accuracy is marked in bold. $\bullet$/$\circ$ indicates that StableMIL is significantly better/worse than the compared methods (paired \textit{t}-tests at 95\% significance level). The last row summarizes the W/T/L counts.}
	\begin{tabular}{ l| l l  l  l | l l  | l }
		\hline
		\multirow{2}{*}{} & \multicolumn{4}{c|}{State-of-the-art multi-instance learning methods} & \multicolumn{2}{c |}{\makecell{Distribution change\\ multi-instance methods}} & \makecell{Proposed \\ method}\\
		\cline{2-8}
		& miSVM & MILES & miGraph & miFV  & MICS  & MIKI & StableMIL \\
		\hline
		% 1 and 2
		gra.os & 56.2 \small{$\pm$ 4.9}$\bullet$ &  61.2 \small{$\pm$ 4.2}$\bullet$  & 68.4 \small{$\pm$ 2.9}$\bullet$  & 61.6 \small{$\pm$ 6.6}$\bullet$ & 70.3 \small{$\pm$ 5.3}$\bullet$ & 68.5 \small{$\pm$ 6.7}$\bullet$& \textbf{71.4 \small{$\pm$ 3.5}}\\ 
		% 1 and 3
		gra.ibm & 57.8 \small{$\pm$ 3.7}$\bullet$ &  57.7 \small{$\pm$ 4.2}$\bullet$  & 64.0 \small{$\pm$ 3.3}$\bullet$  & 57.8 \small{$\pm$ 4.7}$\bullet$ &  62.7 \small{$\pm$ 4.0}$\bullet$  & 65.7 \small{$\pm$ 5.3}$\bullet$ & \textbf{68.6 \small{$\pm$ 4.0}}\\
		% 2 and 3
		mac.win & 56.6 \small{$\pm$ 5.0}$\bullet$ &  61.1 \small{$\pm$ 3.8}$\bullet$ & 67.5 \small{$\pm$ 3.5}$\bullet$ & 67.7 \small{$\pm$ 6.5}$\bullet$ & 68.0 \small{$\pm$ 3.0}$\bullet$ &  69.7 \small{$\pm$ 4.8}$\bullet$ & \textbf{77.2 \small{$\pm$ 2.3}}\\
		% 2 and 4
		os.mac & 54.7 \small{$\pm$ 4.9}$\bullet$ &  59.7 \small{$\pm$ 3.2}$\bullet$ & 61.7 \small{$\pm$ 3.5}$\bullet$ & 58.0 \small{$\pm$ 4.8}$\bullet$ &   62.8 \small{$\pm$ 3.6}$\bullet$  &  \textbf{65.8 \small{$\pm$ 4.3}}$\circ$ & 64.8 \small{$\pm$ 3.7}\\
		% 2 and 5		
		os.win & 63.8 \small{$\pm$ 6.0}$\bullet$ & 64.6 \small{$\pm$ 6.3}$\bullet$ & \textbf{74.7 \small{$\pm$ 3.2}}$\circ$ & 61.8 \small{$\pm$ 7.3}$\bullet$ &   70.2 \small{$\pm$ 2.8}$\bullet$ &  72.2 \small{$\pm$ 5.6} & 72.0 \small{$\pm$ 3.2}\\
		% 7 and 9
		auto.baseball & 54.8 \small{$\pm$ 4.6}$\bullet$ &  60.8 \small{$\pm$ 4.2}$\bullet$  & 55.3 \small{$\pm$ 4.5}$\bullet$  & 59.6 \small{$\pm$ 4.0}$\bullet$ &  58.5 \small{$\pm$ 4.9}$\bullet$ &  \textbf{64.0 \small{$\pm$ 4.5}}$\circ$ & 62.8 \small{$\pm$ 4.0}\\
		% 7 and 8
		auto.moto & 54.5 \small{$\pm$ 5.0}$\bullet$ & 65.5 \small{$\pm$ 7.5}$\bullet$  & 59.5 \small{$\pm$ 3.4}$\bullet$  & 61.7 \small{$\pm$ 5.2}$\bullet$ &  60.4 \small{$\pm$ 4.2}$\bullet$  &  65.6 \small{$\pm$ 6.0}$\bullet$ & \textbf{68.3 \small{$\pm$ 3.6}}\\
		% 9 and 10
		baseball.hockey & 69.2 \small{$\pm$ 8.0}$\bullet$ & 69.6 \small{$\pm$ 3.6}$\bullet$ & 74.1 \small{$\pm$ 3.7}  & 72.2 \small{$\pm$ 6.1}$\bullet$ &   73.2 \small{$\pm$ 3.6} &  \textbf{75.4 \small{$\pm$ 5.7}}$\circ$ & 73.6 \small{$\pm$ 6.0}\\
		% 8 and 9
		moto.baseball & 54.2 \small{$\pm$ 3.8}$\bullet$ & 56.8 \small{$\pm$ 3.9}$\bullet$ & 50.4 \small{$\pm$ 3.4}$\bullet$ & 56.0 \small{$\pm$ 4.1}$\bullet$ & 55.3 \small{$\pm$ 3.1}$\bullet$  & 58.2 \small{$\pm$ 6.9}$\bullet$ &  \textbf{61.2 \small{$\pm$ 2.9}}\\
		% 8 and 10
		moto.hockey & 54.4 \small{$\pm$ 4.5}$\bullet$ &  59.2 \small{$\pm$ 2.7} & 52.5 \small{$\pm$ 3.2}$\bullet$  & 58.0 \small{$\pm$ 4.6}$\bullet$ &   58.0 \small{$\pm$ 3.2}$\bullet$  &  60.1 \small{$\pm$ 4.7} & 60.0 \small{$\pm$ 3.0}\\	
		% 11 and 12		omitted by error
		sci.crypt.elec & 56.5 \small{$\pm$ 8.6}$\bullet$ &  56.5 \small{$\pm$ 5.2}$\bullet$& $57.7$ \small{$\pm$ 4.8}$\bullet$ & 54.5 \small{$\pm$ 4.0}$\bullet$& 60.0 \small{$\pm$ 3.0}$\bullet$ &  61.4 \small{$\pm$ 3.5}$\bullet$&  \textbf{63.4 \small{$\pm$ 2.9}}\\	
		% 11 and 13
		crypt.med & 52.0 \small{$\pm$ 3.9}$\bullet$ & 57.1 \small{$\pm$ 4.1}$\bullet$ & 52.0 \small{$\pm$ 3.8}$\bullet$  & 55.3 \small{$\pm$ 4.6}$\bullet$ & 55.6 \small{$\pm$ 4.0}$\bullet$  &  61.8 \small{$\pm$ 5.7}$\bullet$ & \textbf{74.5 \small{$\pm$ 5.7}}\\	
		% 11 and 14
		crypt.space & 54.8 \small{$\pm$ 6.5}$\bullet$ & 58.1 \small{$\pm$ 2.7}$\bullet$ & 58.2 \small{$\pm$ 1.8}$\bullet$  & 57.3 \small{$\pm$ 5.0}$\bullet$  & 59.2 \small{$\pm$ 4.3}$\bullet$  & 66.1 \small{$\pm$ 5.6}$\bullet$ & \textbf{69.4 \small{$\pm$ 3.4}} \\			
		% 12 and 14
		elec.space & 51.6 \small{$\pm$ 2.9}$\bullet$ & 57.4 \small{$\pm$ 4.2}$\bullet$ & 49.8 \small{$\pm$ 3.6}$\bullet$  & 54.5 \small{$\pm$ 3.7}$\bullet$ & 52.3 \small{$\pm$ 3.0}$\bullet$  &  \textbf{61.9 \small{$\pm$ 6.8}}$\circ$ &  59.8 \small{$\pm$ 4.0}\\			
		% 13 and 14
		med.space & 52.1 \small{$\pm$ 4.0}$\bullet$ & 57.2 \small{$\pm$ 4.1}$\bullet$ & 49.7 \small{$\pm$ 2.7}$\bullet$  & 54.7 \small{$\pm$ 3.8}$\bullet$ &  54.3 \small{$\pm$ 4.0}$\bullet$ &  58.4 \small{$\pm$ 9.4}$\bullet$ &  \textbf{58.8 \small{$\pm$ 6.5}}\\	
		% 16 and 17
		guns.mideast & 58.3 \small{$\pm$ 7.8}$\bullet$ &  54.6 \small{$\pm$ 2.1}$\bullet$  & 59.1 \small{$\pm$ 3.6}$\bullet$ & 58.3 \small{$\pm$ 4.9}$\bullet$ & 60.0 \small{$\pm$ 3.1}$\bullet$  &  60.8 \small{$\pm$ 5.4}$\bullet$ & \textbf{62.1 \small{$\pm$ 3.3}} \\	
		% 16 and 18
		guns.misc & 55.6 \small{$\pm$ 6.8}$\bullet$ & 58.7 \small{$\pm$ 3.5}$\bullet$ & 60.0 \small{$\pm$ 4.2}$\bullet$ & 58.5 \small{$\pm$ 5.8}$\bullet$ & 60.3 \small{$\pm$ 3.9}$\bullet$  &  62.5 \small{$\pm$ 7.1}$\bullet$ &  \textbf{64.4 \small{$\pm$ 2.2}}\\	
		% 17 and 18
		mideast.misc & 55.7 \small{$\pm$ 3.2}$\bullet$ & 57.6 \small{$\pm$ 2.9}$\bullet$ & 63.6 \small{$\pm$ 3.2}$\bullet$  & 58.2 \small{$\pm$ 5.3}$\bullet$ &  62.6 \small{$\pm$ 3.7}$\bullet$  & \textbf{66.3 \small{$\pm$ 4.7}}$\circ$ &  65.7 \small{$\pm$ 2.2}\\	
		% 17 and 19
		mideast.religion & 58.0 \small{$\pm$ 5.8}$\bullet$ &  56.5 \small{$\pm$ 5.1}$\bullet$ & 61.2 \small{$\pm$ 3.8} & 60.0 \small{$\pm$ 5.6}$\bullet$ &  61.6 \small{$\pm$ 3.1} & \textbf{64.4 \small{$\pm$ 7.0}}$\circ$ & 61.5 \small {$\pm$ 3.8} \\	
		% 16 and 19
		religion.guns & 53.5 \small{$\pm$ 3.9}$\bullet$ & 56.7 \small{$\pm$ 2.4}$\bullet$ & 60.4 \small{$\pm$ 6.1}  & 58.0 \small{$\pm$ 5.6}$\bullet$ & 58.6 \small{$\pm$ 3.6}$\bullet$  & 58.7 \small{$\pm$ 3.5}$\bullet$ & \textbf{60.9 \small{$\pm$ 3.9}}\\			
		\hline
		W/T/L & 20/0/0 & 19/1/0 & 16/3/1 & 20/0/0 & 18/2/0 & 12/2/6\\
		\hline
	\end{tabular}
	\label{text}
\end{table*}

\subsection{20 Newsgroups Dataset}
To evaluate the performance of StableMIL for text categorization tasks, we utilize the \textit{20 Newsgroups corpus} \cite{Zhou2009}. It contains paragraphs belonging to 20 different news topics. We use each paragraph as an instance and construct each multi-instance bag with an average of 20 instances. The feature for each instance is represented by the top 200 TF-IDF features. A bag is labeled as positive if it contains a specific topic (i.e., ``graphics''), we use a semantically similar topic as an analogy of the ``cat'' negative concept (i.e., ``graphics'' and ``operating system''), and separate the rest of the topics into two groups to use them as the other negative concepts.
Again, the biased sampling procedure is performed similarly to the last two experiments.
The reported results are averaged over repeating the biased sampling procedure for 30 times.

%The bag construction and biased sampling procedures are performed similarly as the previous section. For example, in the ``gra.ibm'' dataset, we use ``graphics'' and ''ibm''as two positive concepts and randomly separate the rest of the news categories into noisy and negative concepts. The biased sampling procedure is performed similarly as previous experiments, and the results reported are averaged over repeating the biased sampling for 30 times.

%We repeat the experiments for 30 times by using the sampling procedure to generating training and test datasets. The results are reported in Table \ref{text}, with the win/tie/loss counts in the last row and the best results and its comparable ones (paired \textit{t}-tests at $95\%$ significance level) bolded. 

The results on the 20 Newsgroup corpus text categorization task are shown in Table \ref{text}. 
%We can see that the trends are similar to those from image classification task. 
StableMIL performs significantly better than all compared state-of-the-art MIL algorithms.
When compared to distribution change MIL algorithms, the results are similar to those on the MNIST dataset: the performance differences of StableMIL and MIKI are not significant (Wilcoxon rank-sum test $p=0.05$). 
%It is worth noting that although in this task the features are not as refined as those in the MNIST dataset, 
\begin{table}
	\centering
	\setlength{\tabcolsep}{3pt}
	\caption{Testing accuracy (\%) on two real-world distributionally biased image classification tasks from the NICO dataset. 
		%		The highest average accuracy is marked in bold.
	}
	\begin{tabular}{ l| l l  l  l | l  l  | l }
		\hline
		
		Dataset & mi-SVM &  MILES & miGraph  & miFV  & MICS & MIKI  &  StableMIL\\
		
		%		& MILES & miGraph & miFV  & MICS  & MIKI & StableMIL \\
		\hline
		
		% 1 and 3
		Season  & 71.0 & 76.2 & 75.3  & 74.5 & 76.0  & \textbf{82.8} & 81.9\\
		
		% 2 and 3
		Location & 66.5 &  73.5 & 72.0 & 73.2  & 74.1 & \textbf{79.8} & 78.0\\
		\hline
	\end{tabular}
	\label{NICO}
\end{table}
\subsection{Real-world Distribution Biased Dataset}
Recently, a large-scale benchmark dataset NICO, designed for benchmarking distribution-biased image classification has been released on ArXiv \cite{He2019}. Using this data, we evaluate StableMIL on two real-world biased image classification tasks. 

The first task is to distinguish dog vs non-dog images where the training and test samples are biased because of the different seasons during which the photos were taken. The training set contains mostly dog images taken in winter with snowy background, and a small portion of dog images taken in summer with grassy background. In the test samples the bias is reversed with significantly more dog on grass images than dog on snow images. Additionally, the negative images in the test samples also contain more images with grass background than images with snow background, and some images of other animals on grass.
The second task is similar to the first one but the training and test sets are biased due to the different locations where the photos are taken. Specifically, the training set contains mostly dog images in urban environment, including in cage, at home and in street, etc., and less dog images in natural, including dogs in river, grass, and snow, etc. The test samples contain more images of dogs in nature environment and less dogs in urban images. 

To extract multi-instance bags from the images, we use a Mask-RCNN pre-trained on Microsoft COCO dataset \cite{Lin2014} to generate the bounding-boxes and the region in each bounding-box is extracted as an instance. Additionally, the backgrounds are also extracted as instances. The feature vectors for the instances are extracted using an unsupervised convolutional auto-encoder.

The results (Table \ref{NICO}) show that the performances of StableMIL and MIKI are comparable. For other compared methods, StableMIL performs significantly better. Although MIKI performs better than StableMIL on these datasets, the critical drawback of MIKI is that it requires accessing the test samples during training whereas StableMIL does not have this requirement.

\begin{table}
	\centering
	\setlength{\tabcolsep}{3pt}
	\caption{Testing accuracy (\%) on benchmark datasets. The highest average accuracy is marked in bold (paired \textit{t}-tests at 95\% significance level).}
	\begin{tabular}{ l | c c c c| c c |c } 
		\hline
		Dataset & miSVM & MILES  & miGraph  & miFV & MICS & MIKI &  StableMIL\\
		\hline
		Musk1  & 87.4  &  84.2 & \textbf{88.9}  & 87.5 & 88.0 & 88.2 & 87.6\\
		Musk2  & 83.6  & 83.8 & \textbf{90.3} & 86.1  & 90.0 & 91.0  & 90.0\\
		Elephant  & 82.0  & \textbf{89.1} &  86.8  & 82.4 & 86.0 & 85.0 & 84.2\\
		Fox  & 58.2  &  \textbf{76.0} & 61.1 & 58.9 & 72.7 & 66.5 & 63.8\\
		Tiger  & 78.9  & \textbf{86.0} & 85.9 & 79.0 & 86.0 & 83.0 & 78.6\\
		\hline
	\end{tabular}
	\label{benchmark}
\end{table}

\subsection{Classic MIL Benchmark Datasets}
StableMIL can be used without prior knowledge of whether distribution change has occurred. We evaluate StableMIL with classic MIL benchmark non-distributional datasets using repeated 10-fold cross-validation. Each experiment has been repeated for 10 times and the average accuracy are reported in Table \ref{benchmark}. 
The performance of StableMIL is similar to the compared state-of-the-art methods. The results indicate that although StableMIL is designed to produce robust predictions across different training and test distributions, it performs competitively to the state-of-the-art when there is no change between the training and test distribution. This is reasonable, since the causal instances are useful for prediction regardless of whether distribution change has occurred.

It is worth noting that StableMIL performs worse than other methods on the ``Tiger'' dataset. 
This is because in StableMIL only the causal instances are used for classification and all other instances are discarded during bag embedding. This is preferable in a distributional biased setting, since relying on the non-causal instances will mislead the classifier as their distributions change between the training and test data. However, if the training and test data are from the same distribution, including correlated and negative instances is beneficial for two reasons: firstly, it encourages the decision boundary to be positioned away from the negative instances; secondly, correlated instances can be used as ``surrogates'' of causal instances for finding the positive concept if the distributions do not change between the training and test samples. To the best of our understanding, this is the reason why StableMIL does not excel in the experiments of Section 4.5 where the results are obtained from repeated cross-validation and the distributions are not biased. 
%Since causal relationships which are stable across different test distributions are also stable in the original training distribution, therefore the results of StableMIL should be similar to other embedding based multi-instance methods.
%Sign-tests and Friedmann-tests in conjunction with Bonferroni-Dunn at 95\% significance level show that there are no significant differences between these algorithms over five benchmark datasets. 

\subsection{Causal Instances Identification}

\begin{figure}[!t]
	\centering
	\subfloat[Image: 6 and 8]{{\includegraphics[width=0.22\textwidth]{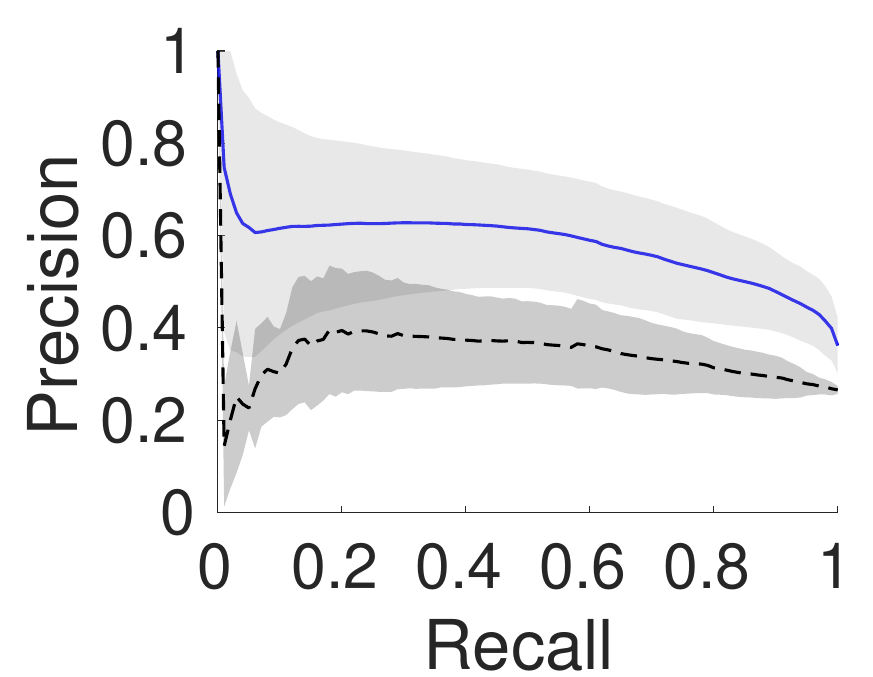}}
		\label{It_Ith}
	}
	\quad
	%	\subfloat[]{{\includegraphics[width=0.22\textwidth]{mac_win.pdf}}}
	%	\quad
	\subfloat[Image: 1 and 7]{{\includegraphics[width=0.22\textwidth]{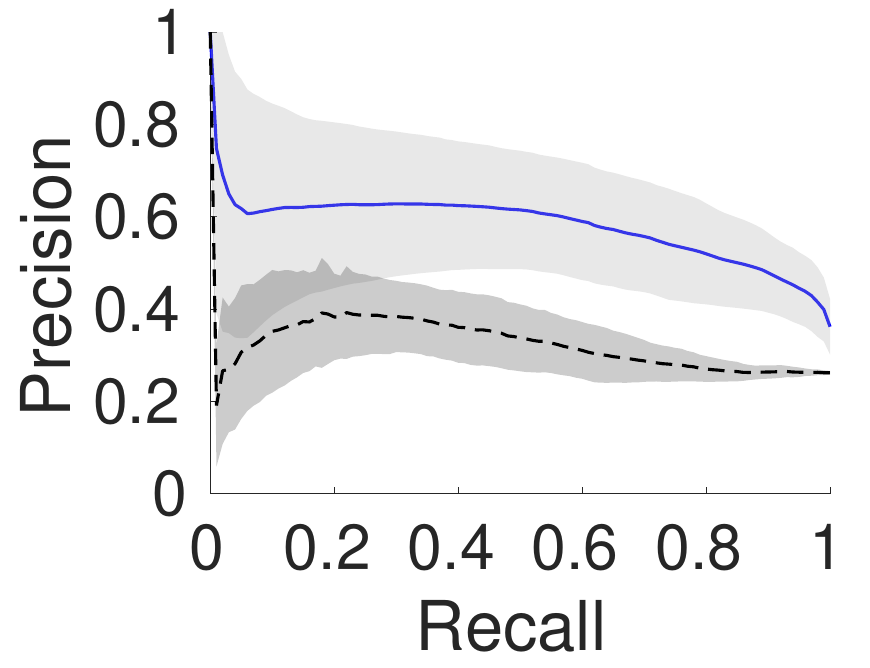}}}
	\quad
	\subfloat[Text: mac.win]{{\includegraphics[width=0.22\textwidth]{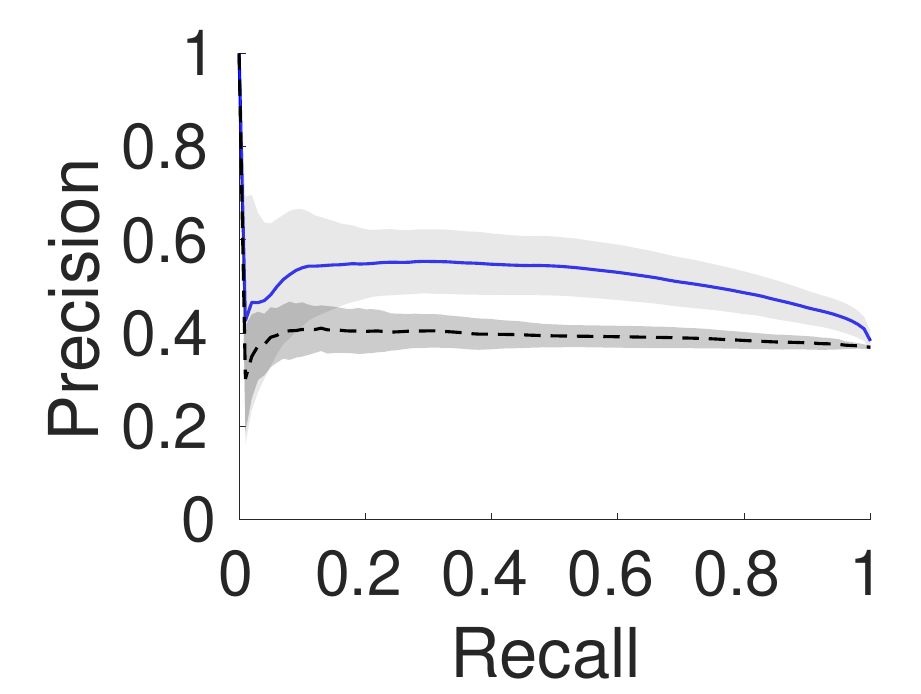}}}
	\quad
	\subfloat[Text: guns.mideast]{{\includegraphics[width=0.22\textwidth]{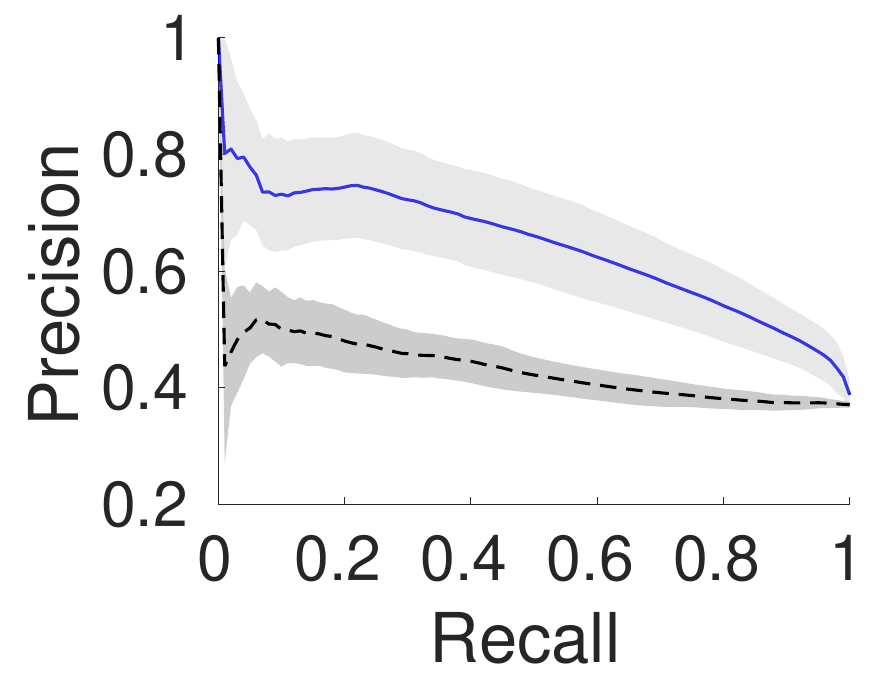}}}
	\quad	
	\caption{The PR curves of causal instances identification (plotted against the stable instance threshold) of StableMIL (blue solid curves) and miVF (black dashed curves). The shaded area indicates the confidence interval.}
	\label{parameter}
	
\end{figure}

We investigate in what degree the stable instances identified by StableMIL are indeed the instances that causes the bag label using both the MNIST and 20 Newsgroup datasets. Since the total number of positive instances and negative instances are usually imbalanced, we use precision and recall instead of accuracy, and report the Precision-Recall (PR) curves for comparison. Figure \ref{parameter} shows the PR-curves (additional figures are included in the supplementary material due to space limit) of StableMIL comparing with miVF \cite{Liu2012}, a MIL algorithm designed for identifying the positive instances in the bags. From the figure we see that StableMIL is effective in identifying the true positive instances, and it consistently outperforms miVF.

\section{Conclusion}
%Most multi-instance learning algorithms assume that the training and test data follow the same distribution, and those methods which deal with distribution change need the test data during the training. 
In this paper, we identify an intrinsic connection between multi-instance learning and causal inference. Inspired by this connection, we propose a novel MIL framework towards robust classification in distributional biased data in without the requirement of accessing the test samples. 
Experiments show that the proposed Stable MIL framework is significantly less sensitive to the distribution changes between the training and test data than existing MIL algorithms, and it performs competitively with state-of-the-art multi-instance distribution change methods \emph{without} the requirement of seeing the unlabeled test data during training.

We have focused on the standard multi-instance assumption and the covariate shift setting in this work. Future work includes investigating the extension of StableMIL to more relaxed distribution change settings, and exploring the link between other MIL assumptions and causal inference.

\paragraph{Acknowledgment} This research is partially funded by Australian Research Council Discovery Project (DP170101306).
\bibliography{reference}

\end{document}